\documentclass[conference]{IEEEtran}
\IEEEoverridecommandlockouts

\usepackage{cite}
\usepackage{amsmath,amssymb,amsfonts}
\usepackage{algorithmic}
\usepackage{multirow}
\usepackage{graphicx}
\usepackage{textcomp}
\usepackage{xcolor}

\def\BibTeX{{\rm B\kern-.05em{\sc i\kern-.025em b}\kern-.08em
    T\kern-.1667em\lower.7ex\hbox{E}\kern-.125emX}}

\usepackage[
    left=0.673in,
    right=0.673in,
    top=0.751in,
    bottom=1.05in
]{geometry}
    
\begin{document}

\title{SGRU: A High-Performance Structured Gated Recurrent Unit for Traffic Flow Prediction
}

\author{

\IEEEauthorblockN{1\textsuperscript{st} Wenfeng Zhang}
\IEEEauthorblockA{\textit{Shandong University} \\
Jinan, China \\
202235331@mail.sdu.edu.cn}
\and
\IEEEauthorblockN{2\textsuperscript{nd} Xin Li*\thanks{* Corresponding Author.}}
\IEEEauthorblockA{\textit{Shandong University} \\
Jinan, China \\
lx@sdu.edu.cn}
\and
\IEEEauthorblockN{3\textsuperscript{rd} Anqi Li}
\IEEEauthorblockA{\textit{Shandong University} \\
Jinan, China \\
2489316785@qq.com}
\and
\IEEEauthorblockN{4\textsuperscript{th} Xiaoting Huang}
\IEEEauthorblockA{\textit{Shandong University} \\
Jinan, China \\
satinhuang@sdu.edu.cn}
\and
\IEEEauthorblockN{5\textsuperscript{th} Ti Wang}
\IEEEauthorblockA{\textit{China Unicom Smart City Research Institute} \\
Beijing, China \\
wangti2@chinaunicom.cn}
\and
\IEEEauthorblockN{6\textsuperscript{th} Honglei Gao}
\IEEEauthorblockA{\textit{Shandong Cultural and Tourism Group Co., Ltd} \\
Jinan, China \\
1348513@qq.com}

}

\maketitle

\begin{abstract}
Traffic flow prediction is an essential task in constructing smart cities and is a typical Multivariate Time Series (MTS) Problem. Recent research has abandoned Gated Recurrent Units (GRU) and utilized dilated convolutions or temporal slicing for feature extraction, and they have the following drawbacks: (1) Dilated convolutions fail to capture the features of adjacent time steps, resulting in the loss of crucial transitional data. (2) The connections within the same temporal slice are strong, while the connections between different temporal slices are too loose. In light of these limitations, we emphasize the importance of analyzing a complete time series repeatedly and the crucial role of GRU in MTS. Therefore, we propose SGRU: Structured Gated Recurrent Units, which involve structured GRU layers and non-linear units, along with multiple layers of time embedding to enhance the model's fitting performance. We evaluate our approach on four publicly available California traffic datasets: PeMS03, PeMS04, PeMS07, and PeMS08 for regression prediction. Experimental results demonstrate that our model outperforms baseline models with average improvements of 11.7\%, 18.6\%, 18.5\%, and 12.0\% respectively.
\end{abstract}


\section{Introduction}
The scenario of multivariate time series occurs in various domains of life. Researchers utilize historical weather data from different regions to predict future rainfall intensity \cite{shi2015convolutional}. Taxi data is used to help cities predict travel resources and reduce traffic congestion \cite{yao2018deep}. Exception monitoring is performed on various states in manufacturing systems and internet services \cite{li2021multivariate}. Since the introduction of Graph Convolutional Networks (GCN) in 2017 \cite{zhang2021semi}, this method has been widely used in the field of Multivariate Time Series (MTS) for spatial semi-supervised and self-supervised learning. By interleaving one-dimensional convolution with gated linear units (GLU) and graph convolution, and appending an output layer after this "sandwich" structure \cite{yu2017spatio}, accurate prediction of traffic flow speed can be achieved. The combination of recurrent neural networks and GCN, with linear layers for output, has made significant contributions to accurate fitting of multivariate traffic flow data \cite{shao2022spatial}. By incorporating attention mechanisms into GCN, it is possible to distinguish the importance of different nodes and utilize Gated Recurrent Units (GRU) for frequency domain feature extraction on long time series, leading to a substantial improvement in prediction accuracy \cite{bai2021a3t}. The aforementioned work demonstrates the effectiveness of GCN in learning spatial features and its strong adaptability to non-Euclidean structured data.
\begin{figure}
  \centering
  \includegraphics[width=8cm,page={1}]{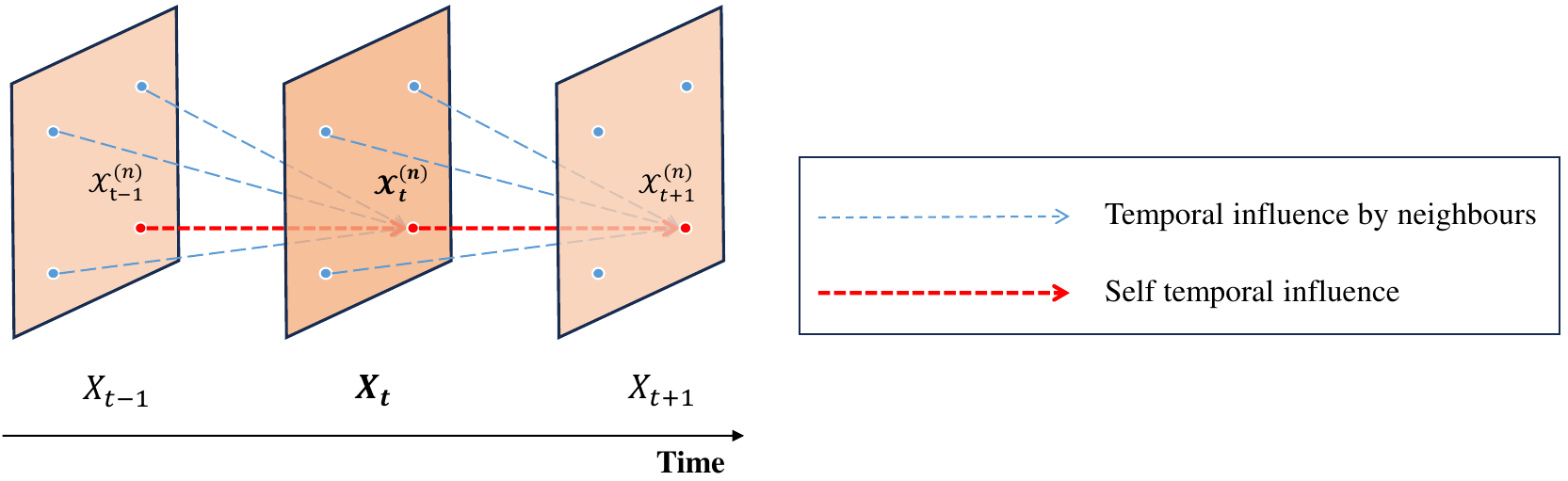}
  \caption{Traffic Flow Prediction: a typical MTS problem}
  \label{fig1}
\end{figure}
\begin{figure}
  \centering
  \includegraphics[width=8cm,page={1}]{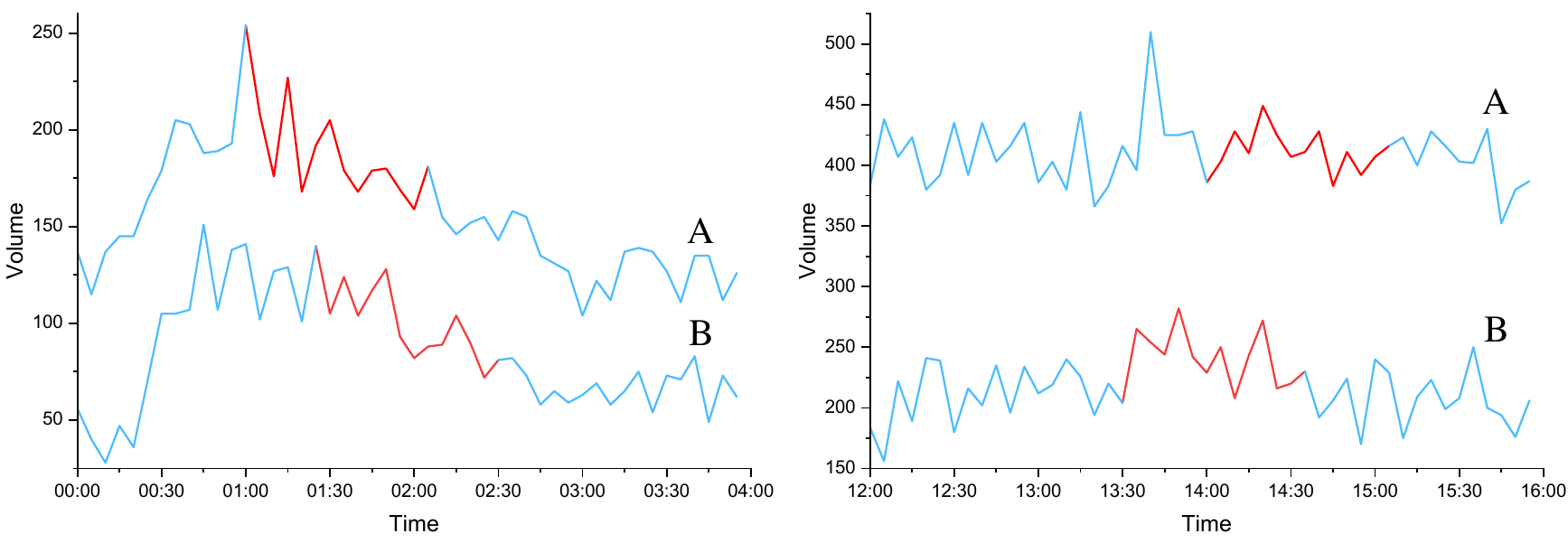}
  \caption{Traffic waveforms of two adjacent nodes (they are similar with a
 lag of about 30 minutes)}
  \label{fig2}
\end{figure}

\subsection{Problem 1: Dilated convolutions break adjacent time steps}
Focusing on the direction of traffic flow prediction (Figure 1,2), recent Temporal Convolutional Networks \cite{bai2018empirical} and Graph WaveNet (GWNet) \cite{wu2019graph} have adopted the mechanism of GCN. However, in the frequency domain, they use dilated convolutions to extract sequence features, which leads to Problem 1: the neglect of the connection between adjacent time steps. The time series in the dataset are discretely sampled, typically every few minutes, and the longer time interval makes the connection between adjacent time steps more important. The use of causal convolutions results in a large number of stacked convolutional layers, increasing the difficulty of training. Additionally, all the convolutional layers share the same convolutional kernel parameters, limiting the features already extracted by the previous layers \cite{ji2022spatio}. Dilated convolutions are used to expand the global receptive field without linearly increasing the depth of convolutional layers and to expedite the convolutional operations. GWNet avoids the vanishing and exploding gradients caused by recurrent neural networks by employing dilated convolutions. The experiments\cite{wu2019graph} select a convolutional kernel size of 2 and an alternating dilation sequence of 1, 2, 1, 2, 1, 2, 1, 2, which achieves the best experimental results. This precisely demonstrates that when applying convolutional operations to the time domain, larger dilations (greater than or equal to 3) should not be used because abandoning crucial information from neighboring time steps leads to a decrease in model performance. Consequently, the experiment employs 8 layers of dilated convolutions to extract sequence features of length 13, significantly increasing the training parameters and time. The traditional dilated convolutions that ignore the connection between adjacent time steps are not a promising approach.

\subsection{Problem 2: Time slices cause a weak of spatial connectivity}
To enhance the connection between adjacent time steps, STSGCN \cite{song2020spatial} utilizes GCN while sampling \textit{T} time steps and treating every 3 time steps as a slice. The sliding window moves only 1 step at a time, generating sequences of length $T-2$. GCN is used to extract spatial features from each time slice, followed by activation functions and linear transformations to restore the original dimensionality. This leads to Problem 2: although the fine-grained time slices largely aggregate the features of adjacent time steps, the lack of inter-slice correlation extraction through GCN results in a weak of spatial connectivity. All the slices are stacked together after GCN, and due to the high repetition in the time dimension, the linear layers require a large number of trainable parameters, extending the model's training time. In order to match each group of 3 time slices with their corresponding adjacency matrix, the trainable parameters of size $N \times N$ are expanded to $3N \times 3N$, nearly exponentially increasing the model size. However, the large number of training parameters is confined within a single slice containing only 3 time steps, greatly limiting the model's temporal receptive field. Therefore, dividing the sequence into time slices is not the optimal choice.

\subsection{Our solution}
To address the aforementioned issues, we propose a Structured Gated Recurrent Units (SGRU) network that treats long time series as a whole. By combining SGRU with Graph Convolutional Networks (GCN) and performing multiple feature extractions on the same sequence, we effectively preserve the correlation information among adjacent time steps. Compared to the linear GRU combined with GCN \cite{bai2020adaptive,zhao2019t}, SGRU utilizes non-linear layers to connect multiple GRUs, thereby enhancing the model's feature extraction capability. Additionally, prior to inputting the data into GRU, we apply multi-layer spatio-temporal embedding. The main contributions of our work are as follows:

\begin{itemize}
\item[$\bullet$] We propose a structured GRU layer that performs more comprehensive feature extraction on temporal data. Compared to a simple linearly connected GRU layer, SGRU has a smaller depth and faster fitting speed.
\item[$\bullet$] We propose a multi-layer spatio-temporal embedding approach that maps high-dimensional features to low-dimensional features and embeds them before the recurrent network. This method enables effective analysis of heterogeneity among nodes.
\item[$\bullet$] We conduct regression predictions on four publicly available California traffic datasets to evaluate the performance of our SGRU model compared to other models. The experimental results demonstrate that our approach generally outperforms the baselines. Additionally, we perform thorough ablation experiments on multiple datasets to verify the effectiveness of the key components of our model. Our code is publicly available on GitHub: https://github.com/atxxfs/SGRU
\end{itemize}

\section{Related Works}
Initially, the prediction of multivariate time series (MTS) was based on statistical methods such as Autoregressive Integrated Moving Average (ARIMA)\cite{williams2003modeling}. Subsequently, with the introduction of Support Vector Machines (SVM)\cite{zhang2011seasonal} by N. Zhang et al. in 2011 and Support Vector Regression (SVR)\cite{chen2015forecasting} by R. Chen et al. in 2015, traditional machine learning theories found their way into the realm of MTS. In recent years, the rapid advancement of deep learning has led to the emergence of numerous deep models focused on feature extraction techniques.
In deep learning approaches, adjacency matrices are employed to represent the relationships between different nodes. Based on whether the adjacency matrix undergoes changes during model updates, it is categorized into static and dynamic adjacency matrices.

\subsection{Static Adjacency Matrices}
Prior to model training, a fixed and unchanging adjacency matrix is generated based on the distances between pairs of nodes. ASTGCN\cite{guo2019attention} integrates three types of attention spans - "hourly", "daily", and "weekly" - to dynamically capture spatiotemporal features using convolution. STGDN\cite{zhang2021traffic} constructs a multi-scale attention network, endowing the model with the capability to capture multi-level temporal dynamics. HGCN\cite{guo2021hierarchical} employs dynamically interacting regional node adjacency matrices to focus on traffic features in the Central Business District (CBD) of a city. FOGS\cite{rao2022fogs} adopts a first-order gradient-supervised method, converting dataset labels from "flow" to "trend" before model training. The benefit of these approaches lies in the interpretability of adjacency matrix values, which align with the fixed geographical distances between observed nodes.

\subsection{Dynamic Adjacency Matrices}
Dynamic adjacency matrices are updated during backpropagation, either through random initialization or pre-definition. In DCRNN\cite{li2017diffusion}, Seq2Seq structures are employed to integrate spatiotemporal features, and recurrent networks are used to predict traffic data. STNN\cite{yang2021space} introduces a novel module to simultaneously capture features in both temporal and spatial domains. ASTTN\cite{feng2022adaptive} separates different time steps of distinct nodes, scaling attention spans to a 1-hop spatial neighborhood to reduce complexity. DSTAGNN\cite{lan2022dstagnn} calculates the proportion of traffic flow on a given day relative to other days using Wasserstein distance, generating spatiotemporal correlation graphs. The advantage of these methods lies in their adaptability to dynamic adjacency changes caused by congestion or unforeseen traffic incidents, aligning more closely with real-world scenarios. Therefore, we also utilize dynamic adjacency matrices.

\section{Preliminary}
In the original sequence
$\mathcal{X} \in \mathbb{R}^{T \times N \times D}$ where $T$ represents the temporal duration, $N$ represents the number of nodes, and $D$ represents the feature dimension, if we denote the current time step as $t$ and sample $\mathcal{X}$ along the temporal dimension, we obtain a sliced tensor. For example, $\mathcal{X}_{t:t + F} \in \mathbb{R}^{(F + 1) \times N \times D}$ where the temporal span is $F+1$. Another example is $X_{t} \in \mathbb{R}^{N \times D}$, which represents the $D$ features of $N$ nodes at the current time step. We define the graph problem of traffic flow as $\mathrm{\mathit{G}}=\left(\mathrm{\mathit{V}}, \mathrm{\mathcal{E}}, \mathrm{\mathcal{A}}, \mathcal{X}_{t-P+1:t}\right)$, where $V$ represents $N$ sensor nodes on the traffic network, and $|V|=N$. $\mathcal{E}$ denotes the edges between the nodes. $\mathcal{A}$ represents the adjacency matrix of $G$, with $\mathcal{A} \in \mathbb{R}^{N \times N}$. $\mathcal{X}_{t-P+1:t} \in \mathbb{R}^{P \times N \times D}$ describes the historical traffic flow observed by $N$ sensors over the past $P$ time steps. Our objective is to learn a prediction function $h(\cdot )$ that predicts the future traffic flow data $\mathcal{Y}_{t+1:t+F} \in \mathbb{R}^{F \times N \times D}$for the next $F$ time steps. This can be expressed as the following formula:
\begin{align*}
\mathcal{X}_{t-P+1: t} \stackrel{h(\cdot)}{\rightarrow} \mathcal{Y}_{t+1: t+F} \tag{1}
\end{align*}

\section{Methodology}
\subsection{Preparation \& Embedding}
\subsubsection{Adaptive graph convolution.}We adopt a method to construct the adaptive adjacency matrix using two spatial embedding matrices $E_{1},E_{2} \in \mathbb{R}^{N \times d}$. The vectors are randomly initialized without any prior knowledge and the parameters are updated during the learning process. The adjacency features of nodes are represented by multiplication:
\begin{align*}
A = SoftMax(ReLU(E_1E_2^T)) \tag{2}
\end{align*}
Two different spatial embedding matrices increase the length of the feature vectors for each node to 2d, in order to improve the model's fitting capability. By multiplying the two embedding matrices, an \textit{N × N} matrix is obtained. The ReLU activation function filters out negative values, and the SoftMax function normalizes the matrix values, resulting in the adaptive adjacency matrix \textit{A}.
\subsubsection{Multilayer spatiotemporal embedding.}  The time embedding vector and spatial embedding vector, $E_{space} \in \mathbb{R}^{N \times d'}$ and $E_{time} \in \mathbb{R}^{d' \times P}$, are randomly initialized. They are added to the original sequence using a broadcasting mechanism and dimension consistency is ensured through linear layers. $W_{e} \in \mathbb{R}^{d \times d'}$ and $b_{e} \in \mathbb{R}^{d'}$ represent the weight and bias of the linear layer, respectively. The introduction of $d'$ is to flexibly change the dimension of the embedding matrix, resulting in ${\widetilde{\mathcal{X}}} \in \mathbb{R}^{P \times N \times d'}$
\begin{align*}
{\widetilde{\mathcal{X}}} = \mathcal{X}W_{e} + b_{e} + E_{space} + E_{time}^{T} \tag{3}
\end{align*}

\subsection{Temporal feature extraction.} To capture the temporal features of the complete time series and prevent gradient vanishing or exploding, we utilize a GRU recurrent network with GCN embedded inside. It is represented as:
\begin{align*}
{LEFT}_{t~} &= A\left\lbrack {{\widetilde{X}}_{t}W_{s}^{(1)} + b_{s}^{(1)},h}_{t - 1} \right\rbrack\\
{RIGHT}_{t~} &= I\left\lbrack {{\widetilde{X}}_{t}W_{s}^{(2)} + b_{s}^{(2)},h}_{t - 1} \right\rbrack \tag{4}\\
{\widetilde{A}}_{t} &= \left\lbrack {{LEFT}_{t},{RIGHT}_{t~}} \right\rbrack W_{a} + b_{a}
\end{align*}
First, we obtain ${\widetilde{X}}_{t}W_{s} + b_{s} \in \mathbb{R}^{N \times H}$ through a linear transformation, where $H$ represents the hidden layer dimension of the recurrent network. $I$ is an $N \times N$ identity matrix, $W_{a} \in \mathbb{R}^{4H \times H}$,$~b_{a} \in \mathbb{R}^{N \times H}$ represent the weights and biases of the linear transformation, respectively. $h_{t-1} \in \mathbb{R}^{N \times H}$ is the output of the previous time step's hidden layer. After obtaining the feature matrix with dimension transformation, denoted as $
{\widetilde{A}}_{t} \in \mathbb{R}^{N \times H}$ , we then input it into the GRU:
\begin{align*}
    z_t &= \sigma(\widetilde{A}_tW_z+b_z)\\
    r_t &= \sigma(\widetilde{A}_tW_r+b_r)\\
    C_{t} &= tanh\left( \left\lbrack {{\widetilde{X}}_{t},{r_{t} \circ h}_{t - 1}} \right\rbrack W_{c} + b_{c} \right) \tag{5}\\
    h_{t} &= \left( 1 - z_{t} \right) \circ C_{t} + z_{t} \circ h_{t - 1}
\end{align*}

\subsection{Structured GRUs network.} A single GRU has limited regression performance. To enhance the model's performance, we adopt a structured multi-GRUs network. Different GRUs extract temporal features of the same sequence, and the hidden layer outputs are aggregated, linearly transformed, and then outputted. The model structure is illustrated in Figure 3, and the comparison between Linear GRUs and SGRU is illustrated in Figure 4.

\begin{figure}
  \centering
  \includegraphics[width=8cm,page={1}]{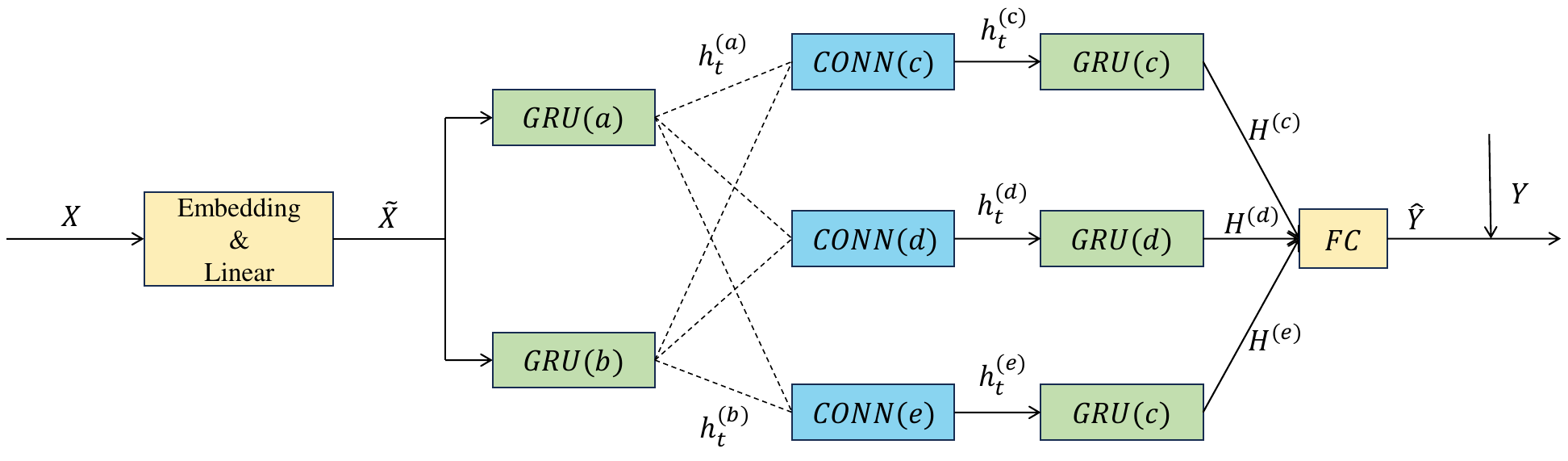}
  \caption{Model structure of SGRU}
  \label{fig3}
\end{figure}

In SGRU, the output of GRU(a) and GRU(b), $h_t^{(a)}$ and $h_t^{(b)}$, respectively, pass through three linear units. $h_t^{(a)}$ additionally undergoes a sigmoid activation and a Hadamard product is applied to connect the two parts as follows:

\begin{align*}
h_t^{(c)} &= \sigma(h_t^{(a)}W_{ac}+b_{ac})\circ(h_t^{(b)}W_{bc}+b_{bc})\\
h_t^{(d)} &= \sigma(h_t^{(a)}W_{ad}+b_{ad})\circ(h_t^{(b)}W_{bd}+b_{bd}) \tag{6}\\
h_t^{(e)} &= \sigma(h_t^{(a)}W_{ae}+b_{ae})\circ(h_t^{(b)}W_{be}+b_{be})
\end{align*}

The output results $h_t^{(c)}$, $h_t^{(d)}$, and $h_t^{(e)} $initialize the hidden layers of GRU(c), GRU(d), and GRU(e), respectively.

The widely used linear GRU (shown in Figure 4, right) operates by feeding the temporal data into the first GRU, passing the output to the next GRU, and using the output of the last GRU for making predictions\cite{zhao2019t}. SGRU improves upon this structure (shown in Figure 3 and Figure 4, left) by introducing the Connection Module (CONN), which simultaneously preserves the nonlinear features from the output of GRU(a) and the linear features from the output of GRU(b), and merges them together. The final prediction includes the outputs of all three SGRUs, resulting in three times the output of the linear GRU and better performances.

\begin{figure}
  \centering
  \includegraphics[width=8cm,page={1}]{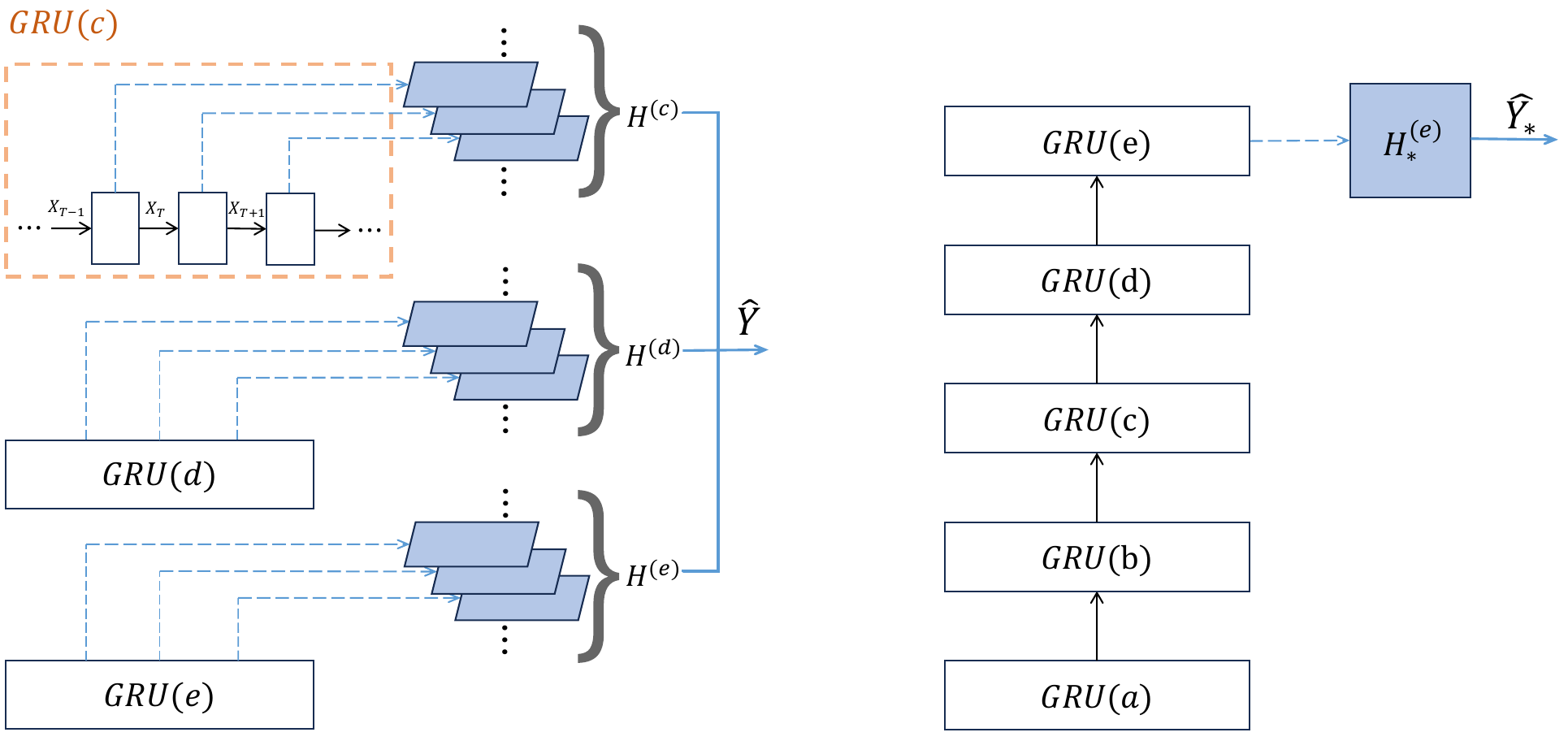}
  \caption{Structural comparison between SGRU (left) and Linear GRUs (right)}
  \label{fig4}
\end{figure}

\subsection{Prediction and loss calculation.} The outputs of the last three GRUs at all time steps, which are $\mathcal{H}^{(c)}$, $\mathcal{H}^{(d)}$ and $\mathcal{H}^{(e)}$, are concatenated and passed through a fully connected (FC) layer to obtain the final result.
\begin{align*}
\mathcal{H} &= \left[h_{t-P+1},h_{t-P+2},...,h_t\right]\\
\hat{Y} &= FC(\left[\mathcal{H}^{(c)},\mathcal{H}^{(d)},\mathcal{H}^{(e)}\right]) \tag{7}
\end{align*}
In order to predict the future states of $N$ nodes over a time horizon of $F$ time steps, The model loss is computed using the mean absolute error (MAE), and the model parameters are updated using back propagation with gradient descent. The Adam optimizer is utilized to minimize the value of $\mathcal{L}$ over multiple iterations.
\begin{align*}
\mathcal{L}(\hat{Y},Y) = \frac{1}{F \times N}\sum_{f=1}^{F}\sum_{n=1}^{N}\left|{\hat{Y}}_f^{(n)}-Y_f^{(n)}\right| \tag{8}
\end{align*}

\section{Experiment}

\begin{table*}[]
\centering
\caption{Performance comparison to other baselines on four publicly available datasets.}
\label{tab:1}
\begin{tabular}{|l|l|lll|lll|lll|lll|}
\hline
Datasets                 &            & \multicolumn{3}{c|}{PeMS03}                                                                & \multicolumn{3}{c|}{PeMS04}                                                                & \multicolumn{3}{c|}{PeMS07}                                                               & \multicolumn{3}{c|}{PeMS08}                                                                \\ \hline
Metrics                  &            & \multicolumn{1}{l|}{MAE}            & \multicolumn{1}{l|}{RMSE}           & MAPE           & \multicolumn{1}{l|}{MAE}            & \multicolumn{1}{l|}{RMSE}           & MAPE           & \multicolumn{1}{l|}{MAE}            & \multicolumn{1}{l|}{RMSE}           & MAPE          & \multicolumn{1}{l|}{MAE}            & \multicolumn{1}{l|}{RMSE}           & MAPE           \\ \hline
\multirow{12}{*}{Models} & DCRNN      & \multicolumn{1}{l|}{18.18}          & \multicolumn{1}{l|}{30.31}          & 18.63          & \multicolumn{1}{l|}{24.70}          & \multicolumn{1}{l|}{38.12}          & 17.12          & \multicolumn{1}{l|}{28.30}          & \multicolumn{1}{l|}{38.58}          & 11.66         & \multicolumn{1}{l|}{17.86}          & \multicolumn{1}{l|}{27.83}          & 11.45          \\ \cline{2-14} 
                         & STGCN      & \multicolumn{1}{l|}{17.49}          & \multicolumn{1}{l|}{30.12}          & 17.15          & \multicolumn{1}{l|}{22.70}          & \multicolumn{1}{l|}{35.55}          & 14.59          & \multicolumn{1}{l|}{25.38}          & \multicolumn{1}{l|}{38.78}          & 11.08         & \multicolumn{1}{l|}{18.02}          & \multicolumn{1}{l|}{27.83}          & 11.40          \\ \cline{2-14} 
                         & GWNet      & \multicolumn{1}{l|}{19.85}          & \multicolumn{1}{l|}{32.94}          & 19.31          & \multicolumn{1}{l|}{25.45}          & \multicolumn{1}{l|}{39.70}          & 17.29          & \multicolumn{1}{l|}{26.85}          & \multicolumn{1}{l|}{42.78}          & 12.12         & \multicolumn{1}{l|}{19.13}          & \multicolumn{1}{l|}{31.05}          & 12.68          \\ \cline{2-14} 
                         & ASTGCN     & \multicolumn{1}{l|}{17.69}          & \multicolumn{1}{l|}{29.66}          & 19.40          & \multicolumn{1}{l|}{22.93}          & \multicolumn{1}{l|}{35.22}          & 16.56          & \multicolumn{1}{l|}{28.05}          & \multicolumn{1}{l|}{42.57}          & 13.92         & \multicolumn{1}{l|}{18.61}          & \multicolumn{1}{l|}{28.16}          & 13.08          \\ \cline{2-14} 
                         & STSGCN     & \multicolumn{1}{l|}{17.48}          & \multicolumn{1}{l|}{29.21}          & 16.78          & \multicolumn{1}{l|}{21.19}          & \multicolumn{1}{l|}{33.65}          & {\underline{13.90}}    & \multicolumn{1}{l|}{24.26}          & \multicolumn{1}{l|}{39.03}          & 10.21         & \multicolumn{1}{l|}{17.13}          & \multicolumn{1}{l|}{26.80}          & 10.96          \\ \cline{2-14} 
                         & STFGNN     & \multicolumn{1}{l|}{16.77}          & \multicolumn{1}{l|}{{\underline{26.28}}}    & 16.30          & \multicolumn{1}{l|}{20.48}          & \multicolumn{1}{l|}{32.51}          & 16.77          & \multicolumn{1}{l|}{23.46}          & \multicolumn{1}{l|}{36.60}          & 9.21          & \multicolumn{1}{l|}{16.94}          & \multicolumn{1}{l|}{26.25}          & 10.60          \\ \cline{2-14} 
                         & STID       & \multicolumn{1}{l|}{19.11}          & \multicolumn{1}{l|}{31.42}          & 18.98          & \multicolumn{1}{l|}{23.94}          & \multicolumn{1}{l|}{37.51}          & 16.12          & \multicolumn{1}{l|}{26.01}          & \multicolumn{1}{l|}{40.60}          & 11.63         & \multicolumn{1}{l|}{19.26}          & \multicolumn{1}{l|}{29.94}          & 11.99          \\ \cline{2-14} 
                         & STGNCDE    & \multicolumn{1}{l|}{19.92}          & \multicolumn{1}{l|}{32.53}          & 19.40          & \multicolumn{1}{l|}{26.35}          & \multicolumn{1}{l|}{41.57}          & 19.90          & \multicolumn{1}{l|}{29.22}          & \multicolumn{1}{l|}{43.79}          & 14.05         & \multicolumn{1}{l|}{19.97}          & \multicolumn{1}{l|}{32.23}          & 13.60          \\ \cline{2-14} 
                         & STSSL      & \multicolumn{1}{l|}{16.77}          & \multicolumn{1}{l|}{26.99}          & 16.28          & \multicolumn{1}{l|}{20.54}          & \multicolumn{1}{l|}{33.14}          & 16.28          & \multicolumn{1}{l|}{22.81}          & \multicolumn{1}{l|}{{\underline{34.71}}}    & 9.95          & \multicolumn{1}{l|}{16.50}          & \multicolumn{1}{l|}{26.51}          & 10.38          \\ \cline{2-14} 
                         & GMAN       & \multicolumn{1}{l|}{{\underline{16.21}}}    & \multicolumn{1}{l|}{26.35}          & {\underline{15.92}}    & \multicolumn{1}{l|}{{\underline{20.23}}}    & \multicolumn{1}{l|}{{\underline{32.17}}}    & 14.29          & \multicolumn{1}{l|}{{\underline{22.12}}}    & \multicolumn{1}{l|}{34.73}          & {\underline{9.58}}    & \multicolumn{1}{l|}{{\underline{16.04}}}    & \multicolumn{1}{l|}{{\underline{25.77}}}    & \textbf{10.08} \\ \cline{2-14} 
                         & StemGNN    & \multicolumn{1}{l|}{17.31}          & \multicolumn{1}{l|}{28.00}          & 17.53          & \multicolumn{1}{l|}{21.06}          & \multicolumn{1}{l|}{33.87}          & 15.91          & \multicolumn{1}{l|}{23.10}          & \multicolumn{1}{l|}{35.67}          & 10.25         & \multicolumn{1}{l|}{17.00}          & \multicolumn{1}{l|}{26.31}          & 11.55          \\ \cline{2-14} 
                         & SGRU(Ours) & \multicolumn{1}{l|}{\textbf{15.22}} & \multicolumn{1}{l|}{\textbf{26.23}} & \textbf{15.64} & \multicolumn{1}{l|}{\textbf{19.96}} & \multicolumn{1}{l|}{\textbf{31.98}} & \textbf{13.22} & \multicolumn{1}{l|}{\textbf{21.60}} & \multicolumn{1}{l|}{\textbf{34.58}} & \textbf{9.16} & \multicolumn{1}{l|}{\textbf{15.96}} & \multicolumn{1}{l|}{\textbf{25.13}} & \underline{10.22}          \\ \hline
                         & Imp.       & \multicolumn{1}{l|}{14.9\%}         & \multicolumn{1}{l|}{10.9\%}         & 11.7\%         & \multicolumn{1}{l|}{12.0\%}         & \multicolumn{1}{l|}{10.5\%}         & 18.6\%         & \multicolumn{1}{l|}{15.0\%}         & \multicolumn{1}{l|}{11.1\%}         & 18.5\%        & \multicolumn{1}{l|}{10.6\%}         & \multicolumn{1}{l|}{10.4\%}         & 12.0\%         \\ \hline
\end{tabular}
\end{table*}

In this chapter, we evaluate the performance of SGRU on publicly available datasets and compare it with other models. Additionally, we conduct ablation experiments by removing certain important modules to verify their effectiveness by comparing the performance with the original model.
\subsection{Datasets}
We evaluate our model on four datasets from the Performance Measure System (PeMS) of the California highway network \cite{chen2001freeway}:
\begin{itemize}
    \item[$\bullet$]\textbf{PeMS03:} 358 nodes, 26,209 time steps, starting from September 1, 2018.
    \item[$\bullet$] \textbf{PeMS04:} 307 nodes, 16,992 time steps, starting from January 1, 2018.
    \item[$\bullet$] \textbf{PeMS07:} 883 nodes, 28,224 time steps, starting from May 1, 2017.
    \item[$\bullet$] \textbf{PeMS08: }170 nodes, 17,856 time steps, starting from July 1, 2016.
\end{itemize}
The time series data for nodes 1, 10 and 20 of PeMS03 are visualized, revealing clear periodicity (left side of Figure 5). Additionally, there are noticeable similarities (before 8 a.m.) and differences (after 8 a.m.) among different nodes within the same day (right side of Figure 5).
\begin{figure}
  \centering
  \includegraphics[width=8cm,page={1}]{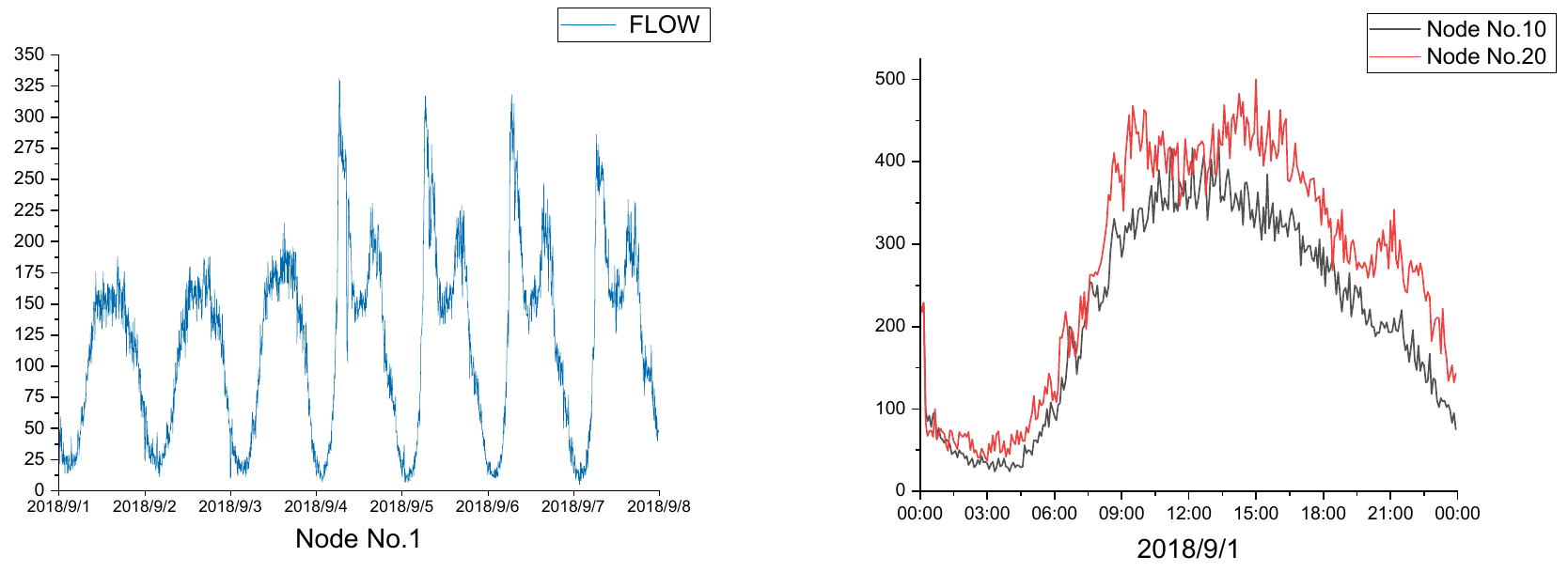}
  \caption{PeMS03 visualization of nodes No.10 and No.20}
  \label{fig5}
\end{figure}\\
Each dataset is preprocessed by linear interpolation to fill missing values, followed by standardization:
\begin{align*}
X' = \frac{X - mean(X)}{std(X)} \tag{9}
\end{align*}
The data is sampled at 5-minute intervals, resulting in 288 time steps per day. The entire dataset was divided into training, validation, and testing sets using a ratio of 6:2:2.
\subsection{Model Settings}
To demonstrate the superiority of our model, we compare it with the following baseline models:
\begin{itemize}
    \item[$\bullet$] \textbf{DCRNN}\cite{li2017diffusion}: Uses a Seq2Seq structure to integrate spatial and temporal features and predict traffic data using recurrent networks.
    \item[$\bullet$] \textbf{STGCN}\cite{yu2017spatio}: Applies graph convolution in the spatial domain and two 1D convolutions in the temporal domain in a sandwich structure.
    \item[$\bullet$] \textbf{GWNet}\cite{wu2019graph}: Utilizes spatially adaptive matrix learning for adjacency transformation and applies dilated convolutions in the temporal domain.
    \item[$\bullet$] \textbf{ASTGCN}\cite{guo2019attention}: Incorporates multiple spatio-temporal attention mechanisms and captures spatio-temporal features using convolution operations.
    \item[$\bullet$] \textbf{STSGCN}\cite{song2020spatial}: Slices long sequences into smaller segments using a sliding time window and enhances local spatial perception on each segment.
    \item[$\bullet$] \textbf{STFGNN}\cite{li2021spatial}: Generates "time graphs" to complement the limitations of "spatial graphs" and learns hidden spatio-temporal dependencies.
    \item[$\bullet$] \textbf{STID}\cite{shao2022spatial}: Applies spatio-temporal embedding to the original sequence, concatenates the embedding results, and outputs them using fully connected layers.
    \item[$\bullet$] \textbf{STGNCDE}\cite{choi2022graph}: The spatiotemporal data is processed using two neural controlled differential equations (NCDEs), and the results are then combined together.
    \item[$\bullet$] \textbf{STSSL}\cite{ji2022spatio}: Learns spatio-temporal heterogeneity through self-supervised mechanisms and combines "heterogeneity-aware enhancement" to improve the model's robustness.
    \item[$\bullet$] \textbf{GMAN}\cite{zheng2020gman}: Uses an encoder-decoder structure with attention mechanisms and incorporates GRU for spatio-temporal feature fusion.
    \item[$\bullet$] \textbf{StemGNN}\cite{cao2020spectral}: Transforms the time domain to the frequency domain using discrete Fourier transform (DFT) and graph Fourier transform (GFT) to capture spatio-temporal dependencies simultaneously.
\end{itemize}
We set the number of historical time steps (P) to 12, the number of predicted time steps (F) to 12, the spatial embedding dimension (d) to 2, the GRU hidden dimension (H) to 64, and the batch size to 64. The experiments were conducted on an \textit{Intel(R) Xeon(R) Gold 6226R} CPU and \textit{NVIDIA GeForce RTX 3090} GPU, and multiple experiments were performed with random seeds ranging from 1 to 10, with the average results being reported.
\subsection{Performance Evaluation}
We selected three evaluation metrics: Mean Absolute Error (MAE), Root Mean Square Error (RMSE), and Mean Absolute Percentage Error (MAPE). Lower values indicate better model performance. The experimental results are presented in Table 1, where we calculated the average values of 11 baseline models across the three evaluation metrics. SGRU consistently achieved improvements of over 10\%, with a particularly significant increase of 18.6\% in MAPE on the PeMS04 dataset. Apart from the MAPE metric on the PeMS08 dataset, where SGRU performed slightly worse than GMAN, SGRU outperformed all other models in terms of all evaluation metrics.

GMAN demonstrated the second-best performance among the models. GMAN incorporates a mechanism of spatial-temporal attention with GRU, and its output results are combined with \textit{L} ST-Attention blocks. However, each block considers the correlation between $t_{j}$ and $t$ time steps, resulting in excessive granularity in the temporal dimension and leading to time step repetition. As a result, the experimental results of GMAN were not as good as those of SGRU.

\subsection{Ablation Study}

To validate the effectiveness of the important modules in SGRU, we introduce the following variants:
\begin{itemize}
    \item [$\bullet$]\textbf{simple:} The original model with 5 linear GRUs and no multi-layer spatiotemporal embedding.
    \item [$\bullet$]\textbf{w/ st-emb:} Adding multi-layer spatiotemporal embedding to the simple model.
    \item [$\bullet$]\textbf{w/ struct:} Replacing the 5 linear GRUs in the simple model with structured GRUs.
\end{itemize}

Due to the large size of the PeMS07 dataset, we conducted experiments sequentially on the PeMS03, PeMS04, and PeMS08 datasets. The evaluation metric was MAE. We varied the value of the future prediction time steps, denoted as F, which corresponded to F=3,6,9,12 (corresponding to 15 minutes, 30 minutes, 45 minutes, and 60 minutes, respectively). The results are shown in Figure 6 and Table 2.

\begin{table}[]
\centering
\caption{MAE of SGRU and its 3 variants}
\label{tab:2}
\resizebox{\columnwidth}{!}{%
\begin{tabular}{|l|l|l|l|l|l|}
\hline
                        &           & 15min & 30min & 45min & 60min \\ \hline
\multirow{4}{*}{PeMS03} & simple    & 14.14 & 15.04 & 15.73 & 16.36 \\ \cline{2-6} 
                        & w/ st-emb & 13.73 & 14.43 & 15.00 & 15.51 \\ \cline{2-6} 
                        & w/ struct & 13.96 & 14.18 & 14.86 & 15.48 \\ \cline{2-6} 
                        & \textbf{sgru}      & \textbf{13.30} & \textbf{14.05} & \textbf{14.67} & \textbf{15.22} \\ \hline
\multirow{4}{*}{PeMS04} & simple    & 18.74 & 19.45 & 20.08 & 20.72 \\ \cline{2-6} 
                        & w/ st-emb & 18.34 & 19.02 & 19.56 & 20.08 \\ \cline{2-6} 
                        & w/ struct & 18.10 & 18.93 & 19.63 & 20.29 \\ \cline{2-6} 
                        & \textbf{sgru}      & \textbf{18.00} & \textbf{18.73} & \textbf{19.35} & \textbf{19.96} \\ \hline
\multirow{4}{*}{PeMS08} & simple    & 15.07 & 15.81 & 16.40 & 17.13 \\ \cline{2-6} 
                        & w/ st-emb & 14.49 & 15.14 & 15.71 & 16.23 \\ \cline{2-6} 
                        & w/ struct & 14.13 & 14.93 & 15.61 & 16.27 \\ \cline{2-6} 
                        & \textbf{sgru}      & \textbf{14.07} & \textbf{14.77} & \textbf{15.38} & \textbf{15.96} \\ \hline
\end{tabular}%
}
\end{table}

\begin{figure*}
  \centering
  \includegraphics[scale=0.6]{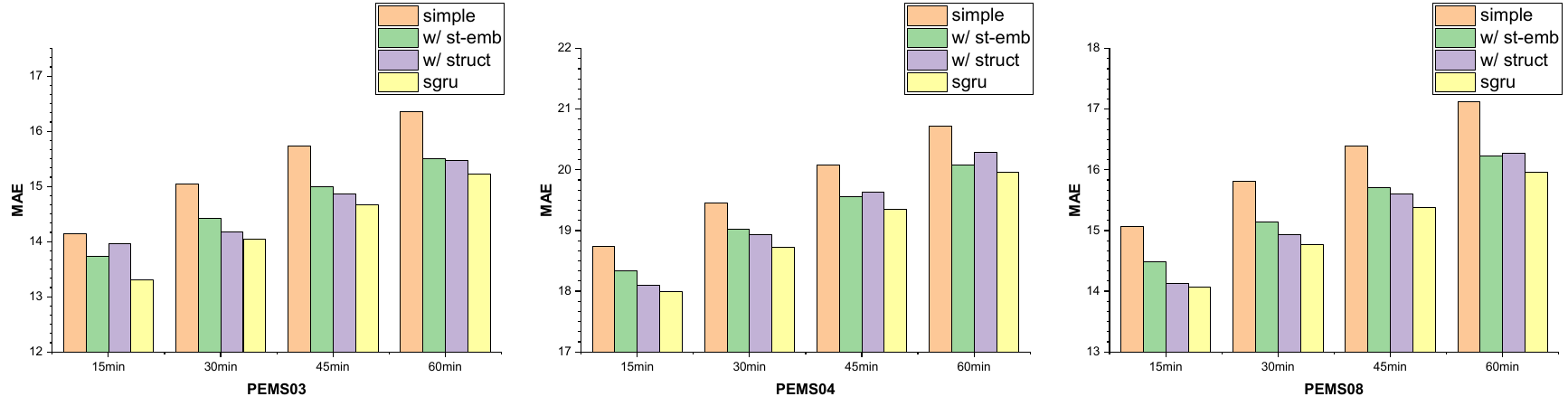}
  \caption{MAE of SGRU and its 3 variants}
  \label{fig1}
\end{figure*}

The simple model performs the worst, the addition of multi-layer spatiotemporal embedding improves its performance, and the use of structured GRUs significantly enhances the performance. SGRU achieves the best performance overall, confirming the contribution of both important modules to the final model.

\section{Conclusion}In order to achieve a better fitting performance on traffic datasets and avoid the loss of adjacent data caused by dilated convolutions or the loose connections between temporal slices, as well as the excessive bulkiness within each slice, we propose a Structured Gated Recurrent Units (SGRU) network. After completing the spatio-temporal embedding, SGRU employs two parallel GRUs to extract features from the same time series. These features are then transformed by non-linear layers and fed into three GRUs. Finally, the hidden layer outputs are concatenated, passed through fully connected layers, and used for regression prediction. We conducted comparative experiments on four publicly available datasets: PeMS03, PeMS04, PeMS07, and PeMS08. Compared to the average performance of baseline models, our approach achieved improvements of 11.7\%, 18.6\%, 18.5\%, and 12.0\% respectively. Furthermore, ablation experiments demonstrated a clear advantage of SGRU over linear GRU.
\section*{Acknowledgment}
This research is sponsored by the National Key R\&D Program of China (2021YFB1407004, 2019YFA0709401) and the Key Program of Shandong Province (2020CXGC010904, 2021SFGC04010401, 2023CXGC010801).

\bibliographystyle{IEEEtran}
\bibliography{IEEEabrv, IEEEexample}
\end{document}